# Spectrum Management for Cognitive Radio based on Genetics Algorithm

Santosh Kumar Singh
Department of Information Technology
Suresh Gyan Vihar University
Jaipur, India
sksmtech@yahoo.com

Gajendra Singh
Department of Computer Sc. and Engineering
Suresh Gyan Vihar University
Jaipur, India
Gajendra881@gmail.com

Vibhakar Pathak
Department of Information Technology
Suresh Gyan Vihar University
Jaipur, India
vibhakarp@rediffmail.com

Dr. Krishna Chandra Roy
Department of Electronics & Communication Engineering
Pacific University
Udaipur, India
Roy.Krishna@rediffmail.com

*Abstract:* Spectrum scarceness is one of the major challenges that the present world is facing. The efficient use of existing licensed spectrum is becoming most critical as growing demand of the radio spectrum. Different researches show that the use of licensed are not utilized inefficiently. It has been also shown that primary user does not use more than 70% of the licensed frequency band most of the time. Many researchers are trying to found the techniques that efficiently utilize the under-utilized licensed spectrum. One of the approaches is the use of "Cognitive Radio". This allows the radio to learn from its environment, changing certain parameters. Based on this knowledge the radio can dynamically exploit the spectrum holes in the licensed band of the spectrum. This paper will focus on the performance of spectrum allocation technique, based on popular meta-heuristics Genetics Algorithm and analyzing the performance of this technique using Mat Lab.

## I. INTRODUCTION

The radio spectrum is a natural resource, and used by transmitters and receivers in a communication network. The allocation of radio spectrum is under control of the central Government; the Federal Communications Commission (FCC) published a report in November 2002, prepared by the Spectrum-Policy Task Force, aimed at improving the way in which this precious resource is managed in the United States [1]. The allocation of the unlicensed frequency bands has resulted in the overcrowding of these bands. The most of the usable electromagnetic spectrum already has been allocated for licensed use, resulting in a shortage of spectrum for new and emerging wireless applications. To resolve this problem, regulators and policy makers are working on new spectrum management strategies. In spectrum reallocation, bandwidth from government and other long-standing users is reassigned to new wireless services such as mobile communications. In spectrum leases, the FCC relaxes the technical and commercial limitations on existing spectrum licenses by permitting existing licensees to use their spectrum flexibly for various services or even lease their spectrum to third parties. The FCC is considering a new spectrum-sharing paradigm, where licensed bands are opened to unlicensed operations on a non-interference basis to licensed operations. Because some licensed bands (such as TV bands) are underutilized, spectrum sharing in fallow sections of these licensed bands can effectively alleviate the spectrum scarcity problem. In this spectrum-sharing paradigm —licensed users are referred to primary users (PU), whereas unlicensed users that access spectrum opportunistically are referred as secondary users (SU) A spectrum hole is a band of frequencies assigned to a primary user, but, at a particular time and specific geographic location, is not utilizing the band. The spectrum utilization can be improved significantly by making it possible for a secondary user (who is not being serviced) to access a spectrum hole on non-interfering basis i.e. SU must quit the bands as arrival of (owner) PU to use spectrum holes The spectrum holes have been utilized to promote the efficient use of the spectrum by exploiting the existence of spectrum holes. This concept is known as Cognitive radio (CR) [2]. A suitable description is found in Haykin paper [3]: "Cognitive radio is an intelligent wireless communication system that is aware of its surrounding environment (i.e., its outside world), and uses the methodology of understanding-by building to learn from the environment and adapt its internal states to statistical variations in the incoming radio frequency (RF) stimuli by making corresponding changes in certain operating parameters (e.g., transmit power, carrier frequency, and modulation strategy, bit-error-rate) in real time. These decisions of parameter changing exploits the objective optimization technique based on different appropriate algorithms. All optimization algorithms are not well performed on decision-making process, however Genetics Algorithms (GA) seems too appealing due to multi-objective optimization capability [4]. The process of decision-making modeled and embedded in Physical [PHY] and medium access control [MAC] layer of conventional wireless communication. The example of CR network architecture is employed in the IEEE 802.22 standard that specifies the air interface ([PHY] and [MAC] layers) for a CR-based wireless regional area network (WRAN) [5].

The paper is structured as follows: In the next section a detailed description of problem formulation and different method for CR spectrum decision-making process is given. In section III a proposed solution and assumption has been made to estimate signal characteristics within the IEEE 802.22 frequency bands are defined for GA. Based on these considerations, evaluation performance of GA is described in section IV. In section V a CR decision-making process results are analyzed by mat lab simulation. Finally, a conclusion is given.

## II. PROBLEM FORMULATION

The aim in case of optimization problems is to get an optimal solution but the difficulty is that the search may get complex and one may not identify where to come across for the solution or where to begin with conventional cognitive approaches derived from rule based systems, expert systems, fuzzy logic, and neural networks. Several methods would find an appropriate solution but the solution might not be the optimum. Each has severe limitations that reduce their operational value in real-time cognitive radio functions, particularly in changing wireless environments. Rule based systems are limited to predetermined capabilities intended into their own rule set. Expert systems are disgracefully weak and dependent on an external expert that must be present when the environmental parameter changes. While fuzzy logic permits approximate solutions to be found in uncertain inputs, the logic has used to locate the approximations does not have a natural evolutionary ability that permits the logic to change in time. The most popular technique for CR modeling is the neural networks, is uncontrollable in that may or may not participate within a set of constraints. Most neural networks involve extensive training to reproduce observed behaviors and usually behave in unexpected ways when problem to solve. The GA inspired cognitive radio model presented in this paper has developed to distribute self-evolution and learning capabilities in terms of the human cognitive development process [5].

### III. PROPOSED SOLUTION AND ASSUMPTION

In this section we first discuss the background of GA and then implementation of proposed solution model on the signal processing of CR.

*A. Background of Genetic Algorithm.*

There are wide applications in fields of science and engineering GA. It has been used to solve complicated problems like, non-deterministic problems and the evolution of simple programs like evolution of images, music and video games. The main advantage of genetic algorithms over the other methods is their multi-objective handling capability. These travels in a search space that uses more individuals for the decision-making and hence are less likely to get fixed in a local extreme like the other existing decision-making techniques. The GA computation starts from the selection of a few randomly generated populations of individuals known as chromosomes that reveal definite characteristics, and follows during the generations. The fitness of each individual in a population in each generation is evaluated based on the crossover and mutation, thus producing a new population, with expects that the population of new generation will be better than the old one. An iterative algorithm is then followed to carry on the process from generation to generation, until convergence criteria met like a maximum number of generations reached or the optimal solution establish and end of the evaluation by the G.A. is described as the Pareto front. [6]. General flow chart for GA has been shown in figure 1.

*B. Implementation of GA in Cognitive Radio*

Our research focuses on the spectrum management with a hypothesis that inputs are provided by either sensing information from the radio environment or the secondary user and they specifies the QoS requirements condition to the radio. The CR senses the radio frequency parameter form the environment at its receiver and engages itself in a decision-making process in order to provide new spectrum allocation as demanded by the user. This requires a decision-making for requirements of SU as parameters like, its modulation method, bandwidth, data rate and power utilization etc. The SU node that needs the spectrum to perform its communications according to QoS requirements, which depends on inputs of SU and CR as gets the information about the RF environment from a sensing components. This permits the decision-making process to make an evaluation between the SU's specifications against the accessible collection of the solutions received from the RF environment. The information that has been sensed from the environment serves as the initial population for the genetic algorithm. We will create random values that will provide as the initial population, which is, assumes as information received from the RF environment and then obtain the decision for allocation as an optimization with the best solution. To maintain the simplicity in the research, the four parameters viz. frequency bands, the modulation method, power and BER will be used. We will use the genetic algorithms that optimized the best function for user QoS requirements from a number of solutions in the pool.

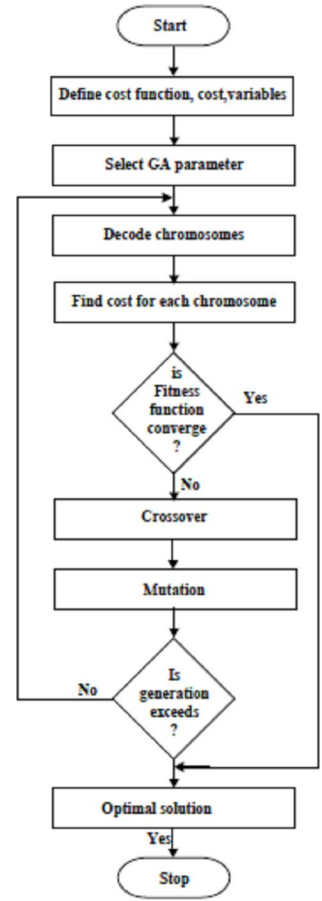

Figure 1. Flow chart for GA

The genetic algorithm approach starts with the definition of the structure of a chromosome. This structure is a sets of genes i.e. frequency, modulation, power and BER in this particular each genes will be considered for the decision-making process fact that a part of the solution. We shall consider the genes from parameters of CR network as IEEE 802.22 discussed in paper [5]. In IEEE 802.22 a CR terminal is allowed to use a radio communication channel with a bandwidth $B(channel)$ = 6 *to* 8 MHz in the frequency range between 41 MHz and 910 MHz so, the overall system bandwidth is $BW$= 869 MHz. This is more than 100 times the signal bandwidth $B$ch. In order to provide a flexible CR system, which is able to optimize the spectral utilization. For the simplicity of frequency gene, we assumed BW=800 MHz and *B(channel)* = 8 MHz frequency band is able to transmit and receive for a particular CR network. The frequency band that we shall consider for this research would band range form 40-840 MHz. with a step size = 8MHz, gives integer values ranging from 0 to 100, as provided by the following table I.

Table I. Frequency gene representation.

| Integer Value | 1 | 2 | …. | 100 |
|---|---|---|---|---|
| Frequency Bands | 40-48 MHz | 48-56 MHz | …. | 832-840 MHz |

One of the important parameters is power gene that we will consider in the chromosome structure definition. This is required by application to give for a high quality transmission, a power less than that may result in poor transmission. The range of power values specified in IEEE802.22 varies from -90dBm to -40dBm. Chromosome structure definition contains the power gene that would be the $2^{nd}$ one in the order ranges from –90dBm to -40dBm with a step size of 1 dBm is characterize in the corresponding integer values shows in the following table II.

Table II. Power gene representation.

| Integer Value | 1 | 2 | …. | 50 |
|---|---|---|---|---|
| Power (dBm) | -90dBm | -89dBm | …. | -40dBm |

Another parameter in chromosome's structure is the bit error rate (BER) defined as radio of erroneous bits to total transmitted bit. BER may vary depends upon certain service applications. The bit error rates can either be reduced by the use of certain modulation and coding schemes at the receiver and transmitter. One more way to reduce the bit error rates is to increase the transmission power of the device that is limited by FCC. In this paper we will consider the bit error rates ranges $10^{-1}$ to $10^{-8}$. This range performs the requirements our applications. The Let us consider step size for the bit error rate is $10^{-1}$. This will give us 8 different values of BER. The following table III shows these values are mapped adjacent to their corresponding BER.

Table III. BER gene representation

| Integer Value | 1 | 2 | …. | 8 |
|---|---|---|---|---|
| BER | $10^{-1}$ | $10^{-2}$ | …. | $10^{-8}$ |

So far one more, and the last of the 4 genes to be considered in structure of definition chromosome is the modulation gene. This paper considers four modulation schemes. These are BPSK, QPSK, 8QAM, and 16QAM so, to represent these four schemes needs 4 integers as shown in table IV.

Table IV. Modulation gene representation

| Integer Value | 1 | 2 | 3 | 4 |
|---|---|---|---|---|
| Modulation | BPSK | QPSK | 8-QAM | 16-QAM |

The four genes as discussed above, all together define structure of chromosome. The arrangement of all genes in the chromosome structure is shown in table V.

Table V. Representation of Chromosome Structure

| Order | 1 | 2 | 3 | 4 |
|---|---|---|---|---|
| Gene | Frequency | Power | BER | Modulation |
| Ranges | 1-100 | 1-50 | 1-8 | 1-4 |
| Bits required | 7 | 6 | 4 | 2 |

As the above table states that the frequency band is the 1st gene in the chromosome structure, contains the any value between 1 and 100. $2^{nd}$ order power gene contains the any value between 1 and 50. BER is the 3rd gene in the chromosome structure, contains any value between of 1 and 8. The last gene in the chromosome structure is the modulation contains any value between 1 and 3. The demands of specific chromosome relative to QoS, from the user or application ask to the CR for an optimal solution to get same structure as that of the chromosome. These four parameters as genes in a chromosome need total 19 bits. This bit string is essential as the mutation operation performs at the bit level. After defining the chromosome structure the next problem is the $1^{st}$ generation population of chromosomes. This can achieve by random generation of $1^{st}$ population of chromosomes within given ranges of each gene defined in table 5. Let us assume that the generation of an initial population size is 50 chromosomes. Further, this population size can be increased up to n numbers, if the outcomes are not acceptable, in the supplied solution set. The fitness in the decision-making process is test over randomly generated 1st population of chromosomes, if results are not satisfactory then goes to a number of operations like selection, crossover and mutation to produce the next generation of fittest chromosomes and process repeats until optimal solution found. However, some of these solutions may be the superior ones while the other ones may be the worst solutions. In order to increase the possibility of good quality solutions in the next generation, the fitness function applied to this initial population and carry out operations like selection, crossover and mutation. Structure of initial population of chromosomes has been shown in table IV.

Table VI. Structure of initial population chromosomes

| Chromosome No | Frequency | Power | BER | Modulation |
|---|---|---|---|---|
| 1 | F1 | P1 | B1 | M1 |
| 2 | F2 | P2 | B2 | M2 |
| …. | …. | …. | …. | …. |
| …. | …. | …. | …. | …. |
| 50 | F50 | P50 | B50 | M50 |
| …. | …. | …. | …. | …. |
| N | Fn | Pn | Bn | Mn |

The definition of a fitness function is required before performing operations like crossover and mutation as shown in figure 1. This fitness function or cost function is fundamentally responsible for the generation of the next generation population chromosomes. This cost function test out for the fitness of the chromosomes in the initial population of chromosomes and passes the fittest of chromosome to the next generation population is discussed in next following section.

## IV. EVALUATION PERFORMANCE OF GENETIC ALGORITHMS

Previously we have generated the first generation population of the chromosomes, now next step is to obtain the fitness evaluation of each chromosome in the population. This 1st population of chromosomes has a different pool of possible solutions that meet the QoS needs of the SU or the application. Some of these solutions may exactly satisfy the QoS requirement while others is just nearby to specifications.

So, choosing amongst this pool to make the trade-off conditions. In this paper, assumption has been made that fitness function that is equally dependent on all the four parameters defined as above as genes and the fitness function uses weighted sum approach in GA. All the four parameters set as an equal weight each. The input is given by SU or application for QoS requirements are compared against the population of chromosomes. As discussed earlier, GA has capability of multi objective optimization, so that overall fitness is computed as the cumulative sum of the individual fitness of each parameter (gene) according to the procedure described as following.

Let parameters $x_1, x_2, x_3$ and $x_4$ be the frequency gene, power gene, BER gene and modulation gene respectively. Therefore these requires fitness function $f_i$ for each parameter given by

$$f_i = \begin{cases} \left[\dfrac{w_i \cdot |x_i - x_i^d|}{x_i^d}\right] & if \ |x_i - x_i^d| < x_i^d \\ w_i & otherwise \end{cases} \quad (1)$$

The overall fitness function value of chromosome $F$ can be calculated as cumulative sum of individual fitness value of all the genes i.e.

$$F = \sum_{i=1}^{4} f_i . \quad (2)$$

The over all fitness value of chromosome in percentages is given by

$$\text{Total fitness value}(\%) = 100 \cdot \left[1 - \sum_{i=1}^{4} f_i\right] \quad (3)$$

Where, $x_i^d$ is desired QoS parameter, $w_i$ is a weight and $\sum_{i=1}^{4} w_i = 1$. where $i = 1, 2, 3$ and $4$ be the gene order of concern parameter as given in table IV. Note that each gene having equal weight i.e. $w = 25\%$, however in true practical case in wireless communication weighting factor $w$ can be vary according to QoS specifications. Also note that using (2), the more the fitness value of the gene, results in the less suitable solution. So, obtaining best solution with the fitness value of the gene converging to zero. On the other hand, using (3) comparison of the total fitness of the chromosomes, the larger the fitness value the better the solution obtained and vice versa. Selection operator of GA engages the selection of the best children within the pool of existing chromosomes. There are two methods uses the selection operator first is "Roulette Wheel selection" method. Probability of selection with each individual best chromosome from population, given by

$$p_i = \dfrac{f_i}{\sum_{i=1}^{n} f_i} \quad (4)$$

Where, $p_i$ is probability of the individual chromosome $f_i$ is the fitness of an individual gene $i$ in the pool and $n$ is the total number of chromosomes in the population. The chromosomes with the maximum probability value transferred to the next generation and not to allow the lower fitness value chromosomes to the next population of chromosomes. Subsequently each generation the fittest chromosomes are transfer to the new population until the new population achieves the maximum given limits.

Second selection method is "Elitism" used in decision-making process that selects the best among the population of chromosomes and transfers to the next generation. In reality, it copies a small number of the best chromosomes and transfers to the new population. It stops the loss of the best possible solution.

The next step is that after the selection of the best chromosomes existing in the pool, to perform the crossover operation. The crossover operation is carrying out on a pair of chromosomes, selected arbitrarily. The operation is completes at the crossover points that define the joint of the genes in the chromosome structure. The crossover operation is shown in figure 2.

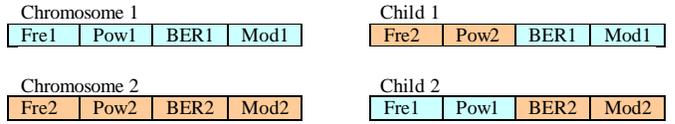

Figure 2. chromosomes crossover representation

Next step is that the mutation operation engages the flip to corresponding single binary digit with single chromosome at a time. In mutation, a randomly selected bit fit in to any of the 4 genes. This provides new solution in search space shown in figure 3.

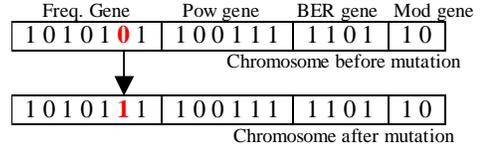

Figure 3. Chromosome mutation representation

The process of all GA operators will repeat until best solution found as shown in flow chart of figure 1. Next section analyses the results with mat lab simulation.

## V. RESULTS

The types of parameters or genes were used in the decision-making process model for user's QoS as:
Frequency band requirement (Freq.) for operation, power requirement (Pow) for operation during the communication, bit error rate (BER) specification, modulation scheme requirement for communication. The following QoS requirement given for testing as: Feq = 50 (i.e 400-408 MHz), Pow = 41 (i.e. 50dBm), BER = 3 (i.e. $10^{-3}$) and Mod = 3 (i.e. 8-QAM), mat lab results observed in following figures as shown below. We have selected population size =20, crossover fraction = 0.8 and number of elite chromosomes =2. (In the case of second method of selection)

For verification of operation the GA was modeled using RF parameters/genes. Several simulations were run and results verified. The output consisted of final optimized gene values. Using (2), it has been verified that the best solution with the fitness value of the gene converging to zero, but in our result show that the converging point is 0.2125 as shown in figure 4. It has also verified that using (3), the larger the fitness value the better the solution obtained and vice versa as shown in figure 5, but our result shows that the best total fitness value (%) turn out to be 78.75%. Lastly, GA found that best value of parameter to obtain required QoS specification for CR as shown in the figure 6.

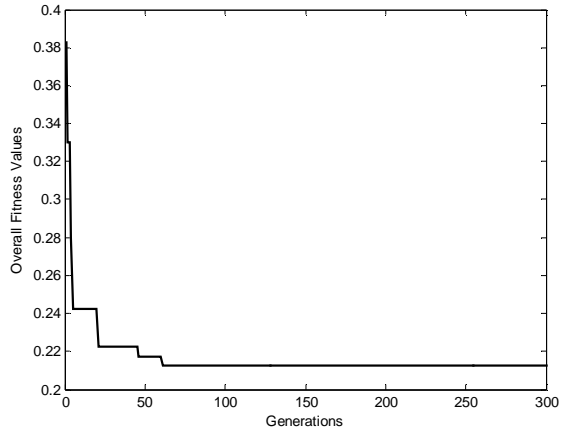

Figure 4. Plot of overall fitness value using (2)

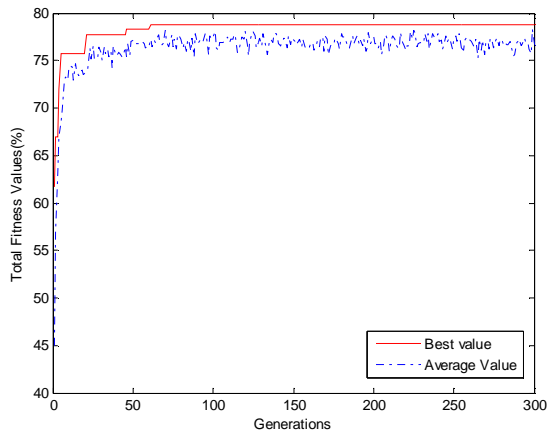

Figure 5. Plot of total fitness value (%) using (3)

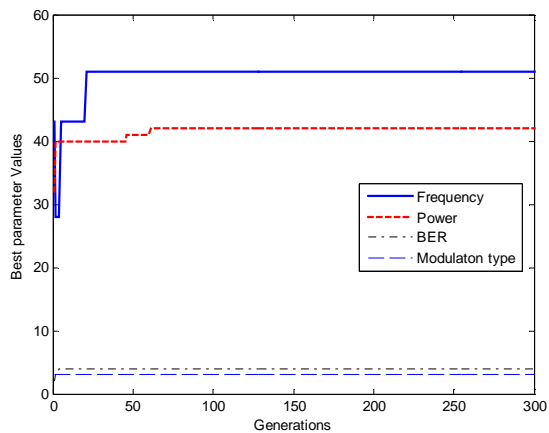

Figure 6. Plot of best parameter values for each gene

## VI. CONCLUSIONS

This work shows that the fitness function of the individual parameters or genes increases with increase in number of generations, but this performance is always not linear. This performance of the G.A. is due to the existence of other genes in the chromosome structure that affect the decision-making process, to reach an optimal solution. This is for the reason that the optimal solution reached by the Genetic Algorithms may have to cooperation for an individual gene to have a better solution for another gene in the structure at the same moment and therefore obtain a better overall fitness value of the chromosomes. G.A actually go for the nearby possible values for each gene along with the available pool of solutions. Also, the range for decision-making connected with each gene affects the decision-making process. A gene with a lesser range i.e. modulation gene in this case have a higher fitness value, while with a bigger range i.e. frequency gene in this case will have a worse fitness value in the optimal solution found by the GA, over the number of generations. This mean that the individual fitness values for the genes may not increase in the same manner, however the total fitness values stay above 80% all through the generations and find the nearby probable best values in the existing pool of solutions

The parameters measured for fitness function in these investigations was independent of each other. Another possible future work of the research can be the consideration other parameters i.e. Noise, channel coding, throughput, delay, packet error loss, free space loss, data-rate, fading and multi path etc. that may consider in the fitness function definition.